# A New Deep State-Space Analysis Framework for Patient Latent State Estimation and Classification from EHR Time Series Data

Aya Nakamura[1], Ryosuke Kojima[1], Yuji Okamoto[1], Eiichiro Uchino[1,2], Yohei Mineharu[1,3], Yohei Harada[1], Mayumi Kamada[1], Manabu Muto[4], Motoko Yanagita[2,5], Yasushi Okuno[1]

1 Department of Biomedical Data Intelligence, Graduate School of Medicine, Kyoto University
2 Department of Nephrology, Graduate School of Medicine, Kyoto University
3 Department of Artificial Intelligence in Healthcare and Medicine, Graduate School of Medicine, Kyoto University
4 Department of Therapeutic Oncology, Graduate School of Medicine, Kyoto University
5 Institute for the Advanced Study of Human Biology (ASHBi), Kyoto University

## Abstract

Many diseases, including cancer and chronic conditions, necessitate extended treatment periods, thereby requiring corresponding long-term treatment strategies. To address this need, there is currently a strong focus on machine learning and AI research aimed at analyzing and learning from electronic health records (EHRs).

Developing effective long-term treatment strategies requires more than merely capturing sequential changes in patient test values; it also entails devising a model capable of capturing the patient's internal state over an extended duration. Moreover, for AI to be applicable in real clinical settings, the model must not only be explainable but also clinically interpretable as a time-series model. However, current AI applications utilizing EHRs have not satisfactorily met these requirements. Researchers have extensively studied time-series models that explicitly learn temporal changes in test data and explainable AIs that identify crucial features. They have also developed models representing internal states like patient pathology as latent states, without considering time series. Nevertheless, none have successfully constructed an explanatory and clinically interpretable model by learning temporal changes in patient latent states from EHRs, which are high-dimensional, large-scale, time-series real-world data.

In this study, we propose a novel framework, the "deep state-space analysis framework," which employs unsupervised learning of EHRs through a deep state-space model, incorporating a time-series frame. This framework enables learning, visualizing, and clustering of temporal changes in patient latent states corresponding to disease progression. To evaluate the framework, we utilized time-series laboratory data from 12,695 cancer patients who underwent anticancer drug treatment. By estimating latent states and identifying key laboratory parameters related to responses to various chemotherapy regimens and prognostic trends, we observed a significant correlation between the estimated latent states and prognosis. We successfully visualized the temporal transition of patient status, from a stable state with a positive chemotherapy response to an unstable state close to death and ultimately to death. Moreover, we identified essential test items during the state transition that were characteristic of each anticancer drug. These results surpass what existing latent state estimation models can provide, as those models do not consider time series, highlighting the usefulness of our framework in constructing time series models of EHRs with interpretability and explanatory properties.

This deep state-space analysis framework is expected to enhance our comprehension of the latent state of disease progression from EHRs, leading to early treatment adjustments, prognostic determinations, and optimal treatment throughout the course of diseases in long-term treatment strategies for cancer and other chronic conditions. Additionally, the application of this framework to various high-dimensional and complex time-series real-world data, beyond EHRs, holds promise for advancing new time-series deep learning techniques.

**Major**

Cancer and chronic diseases often require long-term treatments, making it essential to develop effective treatment strategies. Analyzing and learning from electronic health record (EHR) have led to significant advancements in machine learning and AI for this purpose [1][2][3].

To effectively utilize AI in real clinical settings, it is crucial for models to not only be interpretable but also capable of clinical interpretation, particularly in the context of time-series data. However, existing AI models using EHR have not fully met these requirements. While many studies have focused on explicit learning of temporal changes in test data and identifying important features using explainable AI, other models have been developed to represent the internal states of patients, such as their disease conditions, as latent variables. Latent state estimation assumes the existence of unobservable states behind observable ones and estimates the latent states from observed data. This approach helps to better understand disease progression and incorporate it into treatment strategies. For instance, it has been applied to estimate latent states relevant to predicting outcomes in intensive care unit patients [4] and analyzing the states of patients with chronic diseases like diabetes and hypertension [5]. Nevertheless, no prior successful examples exist where the time-varying latent states of patients have been learned from high-dimensional, large-scale time-series data like EHR, while simultaneously constructing an interpretable and clinically relevant model.

To address these challenges, our research utilizes "deep state-space model"[6], a deep learning-based time-series model that considers state-space representations. We developed a novel framework called the "Deep state-space analysis framework," which utilizes the deep state-space model to learn the time-varying changes in latent states related to disease progression from EHR. To achieve a cancer-type-agnostic analysis of high-dimensional and time-varying data from EHR, a powerful time-series model capable of capturing such variations is needed. The deep state-space model was selected as a suitable deep learning technique to capture time-series changes. As shown in Figure 1, our framework combines the estimation of latent states using the deep state-space model with clustering and visualization modules, enabling the interpretation of latent states. We applied this approach to EHR to estimate latent states relevant to the prognosis of cancer patients undergoing chemotherapy and aimed to identify important test items associated with prognosis.

To evaluate our framework, we focused on chemotherapy with anticancer drugs and applied the deep state-space analysis framework to EHR from Kyoto University Hospital. This allowed us to determine if our framework could adequately capture the responsiveness and adverse effects of chemotherapy and the changes in prognosis. Given the diversity of chemotherapy effects concerning timing and individual differences, accurately understanding the state transitions leading to death is beneficial for adjusting individual treatment plans, making it valuable from the perspective of precision medicine. Furthermore, by analyzing the transitions leading to death or near-death related to cancer treatment, it becomes possible to identify the test items crucial during those critical state transitions.

The latent state of a specific patient at a certain test time is represented as a point in a multidimensional space. We defined the state at the final test time of the time-series data as the ultimate latent state and associated it with the patient's prognosis for interpretation. As a result, we confirmed that our framework can identify states characterizing time-varying latent states more clinically significantly than existing dimension reduction methods. Specifically, the estimated time-varying changes in latent states of patients exhibited a high correlation with prognosis. Successfully visualizing the time-varying transitions of patient states from a stable state where chemotherapy was effective to an unstable state close to death and eventually to death demonstrated the capability of our framework. Moreover, conducting analyses based on types of anticancer drugs allowed us to identify crucial test items at specific state transitions unique to each drug. These achievements were not possible with conventional latent state estimation models that do not consider time-series data. Our framework proves to be a useful method for building interpretable models with both interpretability and explanatory power in constructing time-series models from EHR.

By using this framework to understand the latent states of disease progression from EHR, it becomes possible to achieve early changes in treatments, prognosis assessments, and optimal treatments in long-term treatment strategies for diseases like cancer and chronic conditions. Furthermore, applying this framework to various other high-dimensional and complex time-series real-world data, beyond EHR, holds potential to contribute to the advancement of time-series deep learning.

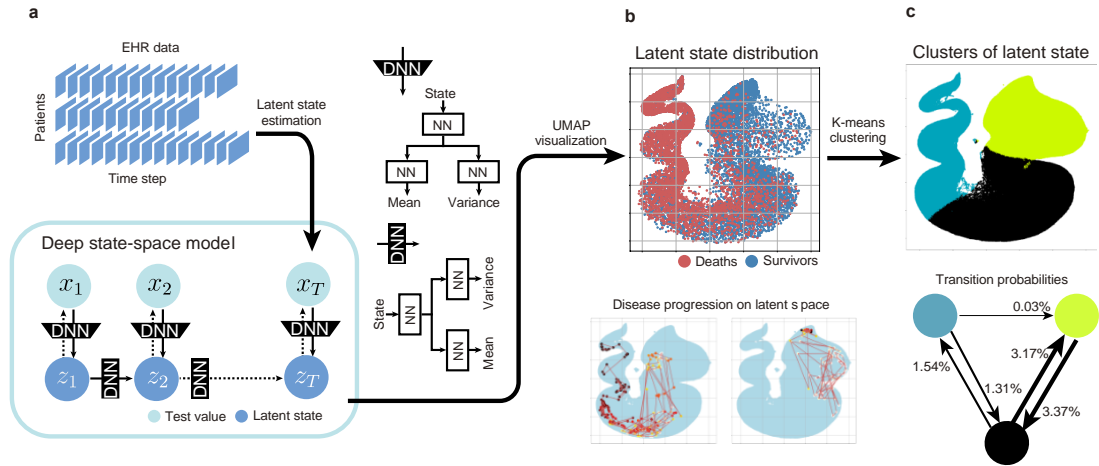

**Figure 1: Overview of our deep state-space analysis framework. a,** Deep state-space models estimate the latent state of electronic health record (EHR) data at each time step to facilitate disease factor analysis. The deep state-space model estimates the latent state of the EHR data at each time step. **b**, The distribution of latent states is visualized by UMAP to clarify the patient's disease progression. **c**, The latent state of patients clustered by k-means clarifies the stage of the disease.

## Results

### Deep state-space analysis framework

The framework consists of latent state estimation of EHR, visualization and clustering of the obtained latent states, using the items of height and weight, vitals, and laboratory tests of EHR, with each patient's item × number of time steps as input. A deep state-space model was used for latent state estimation. deep state-space model is a deep-learning time-series-aware model, which enables the extraction of information over time from large and complex time-series data. To visualize the latent state obtained by multiple dimensions into two dimensions, we used UMAP (Uniform Manifold Approximation and Projection) [7], which is computationally fast, allows efficient dimensionality reduction for large-scale data, and preserves the local and global structure of the data. Clustering was used to stratify the obtained latent states and to associate each state with a poor prognosis group for interpretation. The k-means method was used in the clustering.

### Dataset characteristics

From the electronic medical record data of Kyoto University Hospital, 17,517 cancer patients who received anticancer therapy between January 1, 2006 and October 31, 2018 were selected. The analysis included 12,695 patients, excluding those with at least one item that was missing in all and those with a series length (time steps) of less than 50.

**Visualization and clustering of latent states**

UMAP was applied to the latent states for the reduction of dimension to two-dimensional, and the visualization results are shown in Figure 2a. The red plots indicate the final state of the time series for dead patients, and the blue ones indicate the final state of the time series for surviving patients. The final state in the time series of dead patients can be interpreted as near death and dangerous, while the final state in the time series of surviving patients can be interpreted as far from death and stable. As shown in Figure 2a, the red and blue plots are separated, indicating that the latent states obtained by the deep state-space model successfully capture clinically interpretable patient states.

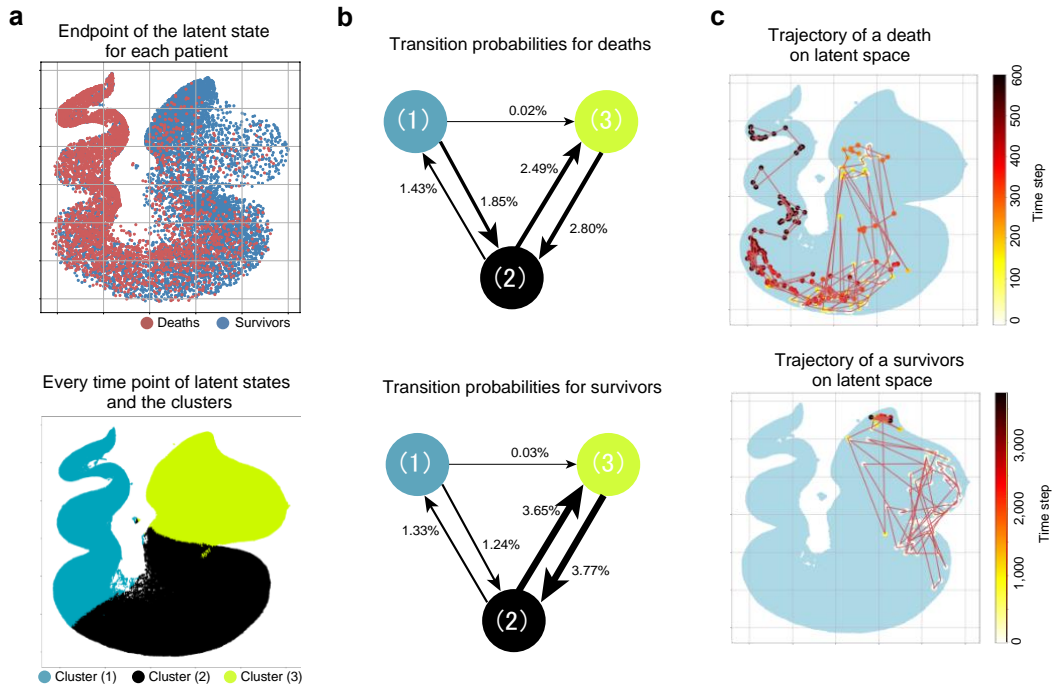

**Figure 2: Properties of latent space of this deep state-space model. a**, Distribution of endpoint and every time point of latent states: utilizing a deep state-space model to infer latent states, which are then reduced to two dimensions using UMAP. The red represents the endpoint of the latent state of death, while the blue represents those survivors. The results of the k-means clustering are shown in blue, black, and yellow. **b**, Inter-cluster transition probabilities for deaths and survivors. **c**, The representative trajectories of the latent state in deaths and survivors.

The results of clustering the latent states at all time points using the k-means method are shown in Figure 2a. The number of plots of the final state of dead and surviving patients in each cluster is shown in Table 1. Cluster (1) can be interpreted as a concentration of patients in a near-death and dangerous state, while cluster (3) is a concentration of patients in a relatively stable state. Cluster (2), with its mixture of red and blue plots, represents an intermediate state.

Table 1: Number of plots of final status of dead and surviving patients in each cluster

|  | Cluster (1) | Cluster (2) | Cluster (3) |
|---|---|---|---|
| Dead patients [persons] | 3205 | 1217 | 246 |
| Survive patients[persons] | 1240 | 3396 | 3391 |

**Latent state transitions**

The number of transitions between clusters for all cancer patients treated with anticancer drugs is shown in expended data 1 Table 1.1 of the supplements. The overall number of transitions was 1,648,039. The number of transitions between clusters of dead and surviving patients is also shown in expended data 1 Table 1.2 and 1.3 of the supplements. The total number of transitions for dead patients was 673,802, while the total number of transitions for surviving patients was 974,237.

For all patients, transitions between clusters were not frequent. Specifically, as shown in Figure 2b, transitions from cluster (1) to (2) and from (2) to (3) were present, but transitions from (1) to (3) were rarely observed. Thus, transitions from (1) ⇄ (2) ⇄ (3) were found to be the predominant pattern. It was also clear that more transitions from (1) to (2) were observed in dead patients and fewer transitions from (2) to (3) were observed in comparison to surviving patients.

The latent state transitions of each patient over time are visualized. Examples of latent state transitions for a dead patient and a surviving patient are shown in Figure 2c. The color bar indicates the number of days elapsed until death, and the color of the plot changes as time passes, from white to red, from red to black, and so on. The light blue plot shows the latent state of all patients over the entire time series. The states are gradually transitioning over time. The dead patients gradually transitioned to a more dangerous state, while the surviving patients tended to remain in a stable state.

**Comparison with conventional methods**

The results of latent state estimation using PCA (principal component analysis), VAE (variational auto-encoder), and a linearized version of the deep state-space model (hereafter simply referred to as the linear state-space model) as a comparison method for the deep state-space model are shown in Figure 3 below.

The top panel of Figure 3 shows the final latent state of the time series, with the red plots representing dead patients and the blue plots representing surviving patients; PCA and VAE did not show sufficient separation between dead and surviving patients, and the distribution of each plot was skewed. On the other hand, the linear state-space model showed a slight separation of the red and blue plots. The most separation was seen in the deep state-space model, where the red and blue plots were spatially separated.

The lower panel of Figure 3 shows the change in the patient's latent state over time; PCA and VAE did not adequately capture the change in state over time. On the other hand, the linear state-space model showed transitions over time, but the deep state-space model was able to capture more detailed transitions. These results indicate that the deep state-space model was able to capture the most mortality and survival states, as well as more detailed state transitions over time.

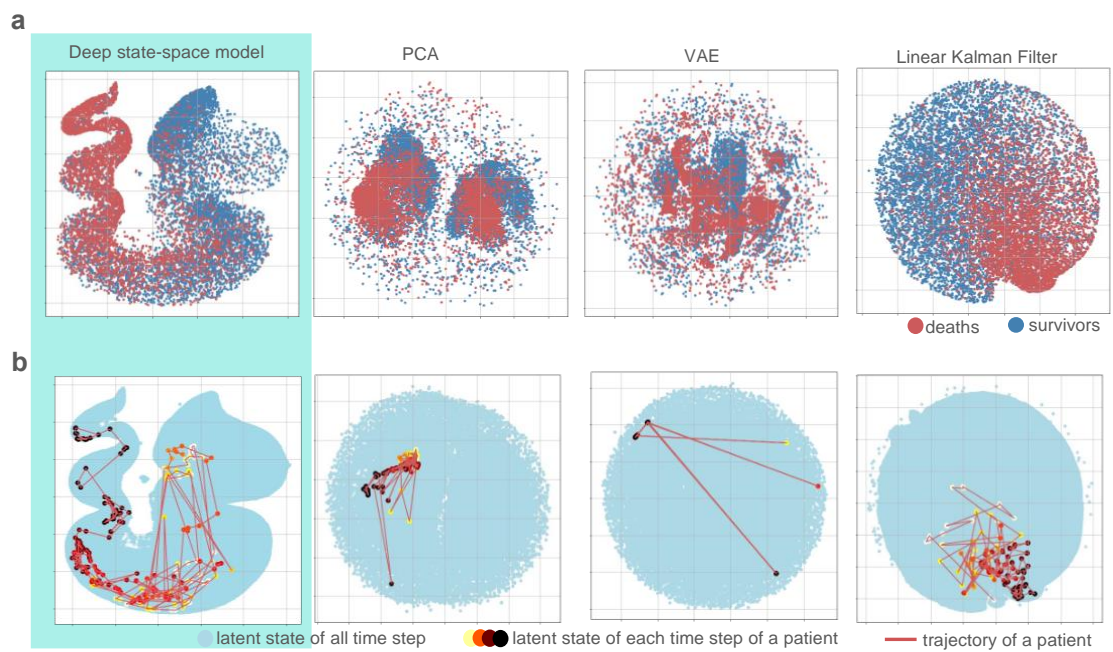

**Figure 3: Comparison with other latent space estimation methods. a**, The distribution of latent state for each method: our method, PCA, VAE, and linear state-space model. Each point is endpoints of latent states and reds and blues are deaths and survivors, respectively. **b**, A trajectory of latent state leading up to death. As leading up to death, the markers at the point of incubation become darker.

### Extract signals during state transitions

Trends in test items values for clusters (1) through (3) were investigated. Figure 4a shows the 10 tests items with large differences between the mean and normal values when the mean value of test items in each cluster was calculated. The horizontal axis represents the mean values of the test items, where the normal value is 0. The values closer to 1 indicate more abnormal high values, and the values closer to -1 indicate more abnormal low values. Test values within the standard values used in the hospital were treated as normal values, those higher than the standard values as abnormal high values, and those lower than the standard values as abnormal low values. Compared to the intermediate state in cluster (2), low values of RBC (red blood cell count), HGB (hemoglobin), and HCT (hematocrit) were observed in the dangerous state in cluster (1). In addition, low values of Ca (calcium), CK (creatine kinase), and TP (total protein) were observed in cluster (3) to cluster (2). Furthermore, in common with clusters (1) to (3), high levels of CRP and D-dimer were observed, and these test items were particularly significantly higher in cluster (1).

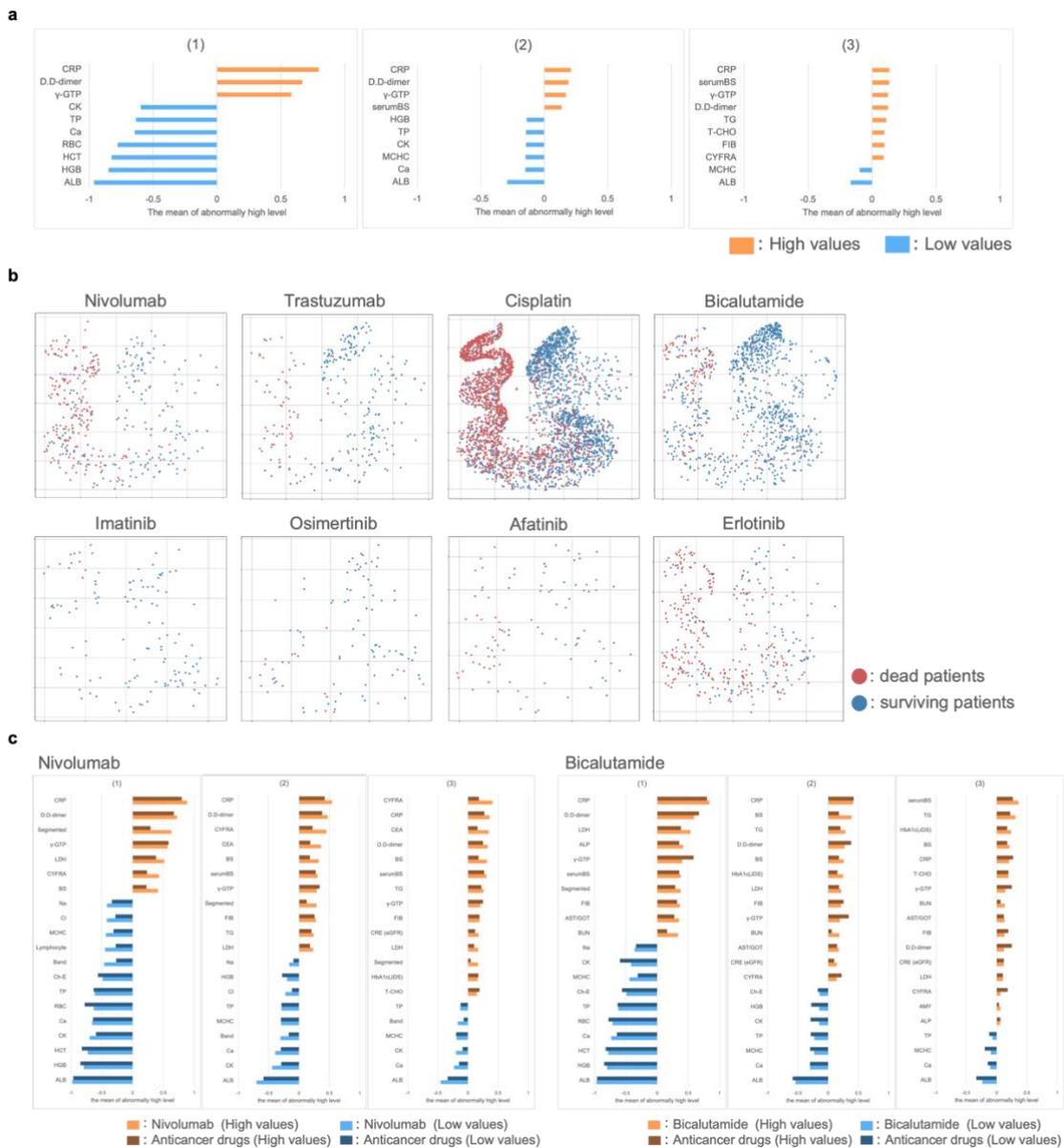

**Figure 4: Test results and drug dependence for each cluster of the latent space. a**, The average degree of abnormality and its main test items in each cluster are shown. The mean abnormality of each test is the average within each cluster of abnormal high (+1), abnormal low (-1), and normal (0)

values for each patient. **b**, The distribution of latent state endpoints for patients treated with each drug. **c**, Mean abnormal values within each cluster specific to nivolumab-treated and bicalutamide-treated cancer patients.

### Visualization and interpretation of the latent state of each injected cancer patient

We visualized and interpreted the latent state for each cancer patient treated with the eight anticancer drugs in Table 2. The analysis was focused on various types of representative anticancer drugs, and test items that are important risk factors during state transitions were extracted.

Table2: List of anticancer drugs

| Drug name | Type |
| --- | --- |
| Nivolumab | Immune checkpoint inhibitors (anti-PD-1 antibodies) |
| Trastuzumab | Molecular targeted drugs (anti-HER2 antibodies) |
| Cisplatin | Platinum drugs |
| Bicalutamide | Hormone therapy drugs |
| Imatinib | Molecularly targeted drugs (tyrosine kinase inhibitors) |
| Osimertinib | Molecularly targeted drugs (tyrosine kinase inhibitors) |
| Afatinib | Molecularly targeted drugs (tyrosine kinase inhibitors) |
| Erlotinib | Molecularly targeted drugs (tyrosine kinase inhibitors) |

The results of visualizing only the latent state of each anticancer patient are shown in Figure 4b. The breakdown of the number of dead and surviving patients for each anticancer administered cancer patient and the number of final state plots for dead and surviving patients in each cluster are shown in Table 2 of expended data 2. The latent state of each cancer patient with cancer treated with each anticancer drug reflected the mortality state as well as the latent state of all cancer patients with cancer treated with all anticancer drugs. In addition, no latent state specific to each cancer patient treated with anticancer drugs could be identified.

The mean values of each test item in clusters (1) to (3) were investigated for cancer patients treated with each anticancer drug, and the 20 test items with the largest differences between the mean and normal values are shown in Figure 1 of expended data 2. Figure 4c is representative of the results for cancer patients treated with nivolumab and ones treated with bicalutamide.

Low values of RBC (red blood cell count), HGB (hemoglobin), and HCT (hematocrit) were important factors also in cancer patients treated with nivolumab and ones treated with bicalutamide, as were state transitions to dangerous states. In addition, high Segmented (neutrophil segmented nuclei) and low lymphocyte counts were observed in cluster (1) as laboratory parameters specific to cancer patients treated with nivolumab. In addition, low levels of CK (creatine kinase) and high levels of LDH (lactate dehydrogenase) and BUN (urea nitrogen) were observed as characteristic signals of transition to a dangerous state in cancer patients treated with bicalutamide.

## Discussion

The latent state obtained by the deep state-space model captures both the death state and the survival state, as shown in Figure 2a, and the change in the state of the dead patient over time to a dangerous state, as shown in Figure 2c. Thus, it can be said that the deep state-space model succeeded in estimating clinically interpretable latent states, capturing both dead and surviving states, and that the obtained latent states captured temporal changes in the patient's state.

The deep state-space model estimated latent states with greater qualitative validity than PCA, a simple linear dimensionality reduction method that does not consider time series, VAE, a nonlinear dimensionality reduction method that does not consider time series, and the linear state-space model, a linear dimensionality reduction method that does consider time series. The deep state-space model is a deep learning type time-series model, and deep learning enables the processing of large amounts of data, making it possible to analyze a wide variety of cancer patients including various cancer types. This deep learning enables the processing of large amounts of data and the analysis of diverse cancer patients, including various types of cancer. In addition, by considering time series, it was possible to capture temporal changes in the patient's condition.

In addition, the study identified risk factors that were important at the time of disease transition for each cancer patient who treated with eight anticancer drugs. The results showed that cancer patients treated with several drugs, including nivolumab and bicalutamide, had characteristic factors that signaled disease progression. Specifically, abnormalities in the immune system, including neutrophils and lymphocytes, were found to be important factors in cancer patients treated with nivolumab. Since nivolumab is a type of immune checkpoint inhibitor, it is reasonable to observe abnormalities in the immune cell system. In previous reports, NLR (neutrophil/lymphocyte count ratio), PLR (platelet/lymphocyte count ratio), and CAR (CRP/albumin ratio) have been considered poor prognostic factors in cancer patients receiving nivolumab in certain cancer types [8][9][10]. In the present analysis, low lymphocyte counts, high CRP counts, and low albumin counts were also observed, consistent with previous reports. Abnormal CK, BUN, and LDH were also observed in cancer patients treated with bicalutamide. Elevated LDH has also been reported in previous meta-analyses and multivariate estimates of poor prognostic factors in prostate cancer patients [11][12]. In the present study, CK and BUN, in addition to LDH, were considered to contribute to the prediction as important laboratory parameters during the transition to a clinically relevant risk state. Also, low levels of HCT, RBC, and HGB are candidates for laboratory tests that are important factors during the transition to at-risk status in cancer patients treated with all anticancer drugs. These abnormal values are indicative of anemia. There are many possible causes of anemia in cancer patients treated with anticancer drugs, including hematopoietic failure, bleeding, bone marrow infiltration, treatment, and nutritional disorders brought on by chronic inflammation caused by the cancer.

A limitation of this study is that this is a retrospective study, making it difficult to determine the causal relationship between these factors. Prospective validation should clarify this.

In the future, we believe that more detailed mechanisms of state transitions can be clarified by investigating which factors are important in the transition to a dangerous state and which causes are responsible. In addition, future work is needed to identify important test items during state transitions for anticancer drugs other than the eight drugs analyzed in this study and for multiple drugs combinations.

## Methods

In this study, latent states were estimated using a deep learning state-space model, the deep state-space model, using patient observations from electronic medical record data. The obtained latent states were then interpreted.

### Ethics review

This study was approved by the Ethics Committee of Kyoto University Hospital (Approval No.: R1498). Given that the data employed in this study were only the existing information obtained from clinical practice, the study was conducted by an opt-out method under Japanese law. We posted an announcement regarding this study on our hospital and department website and provided information on exclusion from participation.

### Dataset

The dataset for this study was created using electronic medical record data from Kyoto University Hospital, a tertiary teaching hospital in Japan, covering the period from January 1, 2006 to October 31, 2018. Eligible patients were cancer patients who were at least 20 years old, had had serum creatinine measured as one of common items for blood tests, and had received at least one orally administered or injectable anticancer drug, excluding patients without missing all items and those with a series length (time steps) of less than 50 steps. A total of 12695 patients were included.

Details of the dataset are presented in Table 3.1. There is overlap in cancer type if the patient has more than one cancer. For cancer type, we used the intermediate classification item groups for neoplasms in ICD-10 shown in Table 3.2.

Table3.1: Composition of target patients

Statistics displayed: Number of patients [persons]

| | | All | Dead | Cancer type (ICD-10 classification) | | | | | | | | | | | | | | | | | |
|---|---|---|---|---|---|---|---|---|---|---|---|---|---|---|---|---|---|---|---|---|---|
| | | | | C00–C14 | C15–C26 | C30–C39 | C40–C41 | C43–C44 | C45–C49 | C50 | C51–C58 | C60–C63 | C64–C68 | C69–C72 | C73–C75 | C76–C80 | C81–C96 | C97 | D00–D09 | D10–D36 | D37–D48 |
| All | | 12695 | 4668 | 41 | 808 | 187 | 0 | 0 | 28 | 3 | 1310 | 2281 | 673 | 94 | 434 | 859 | 7 | 0 | 8 | 84 | 26 |
| Dead | | 4688 | — | 13 | 310 | 53 | 0 | 0 | 10 | 2 | 401 | 598 | 208 | 30 | 112 | 414 | 2 | 0 | 2 | 22 | 7 |
| gender | Male | 7180 | 2867 | 29 | 455 | 121 | 0 | 0 | 20 | 2 | 71 | 2258 | 475 | 65 | 163 | 466 | 3 | 0 | 5 | 8 | 19 |
| | Female | 5515 | 1801 | 12 | 353 | 66 | 0 | 0 | 8 | 1 | 1239 | 23 | 198 | 29 | 271 | 393 | 4 | 0 | 3 | 76 | 7 |
| age-group (years old) | 20–39 | 561 | 185 | 1 | 18 | 8 | 0 | 0 | 1 | 0 | 92 | 6 | 8 | 3 | 5 | 54 | 1 | 0 | 0 | 10 | 0 |
| | 40–59 | 2673 | 1003 | 10 | 179 | 34 | 0 | 0 | 7 | 2 | 479 | 121 | 75 | 11 | 81 | 212 | 3 | 0 | 1 | 34 | 2 |
| | 60–79 | 7792 | 3078 | 25 | 511 | 119 | 0 | 0 | 18 | 1 | 634 | 1448 | 421 | 66 | 283 | 511 | 2 | 0 | 7 | 33 | 22 |
| | 80– | 1669 | 402 | 5 | 100 | 26 | 0 | 0 | 2 | 0 | 105 | 706 | 169 | 14 | 65 | 82 | 1 | 0 | 0 | 7 | 2 |

Table3.2: ICD-10 classification

| C00–C14 | Malignant neoplasms of the lips, mouth and pharynx |
|---|---|
| C15–C26 | Malignant Neoplasms of the Digestive Organs |
| C30–C39 | Malignant neoplasms of respiratory and intrathoracic organs |
| C40–C41 | Malignant neoplasms of bone and articular cartilage |
| C43–C44 | Malignant Neoplasms of the Skin |
| C45–C49 | Malignant neoplasms of mesothelium and soft tissue |
| C50 | Malignant Neoplasms of the Breast |
| C51–C58 | Malignant Neoplasms of the Female Genital Organs |
| C60–C63 | Malignant neoplasms of the male genitalia |
| C64–C68 | Malignant neoplasms of the renal urinary tract |
| C69–C72 | Malignant neoplasms of the eye, brain, and other parts of the central nervous system |
| C73–C75 | Malignant neoplasms of the thyroid gland and other endocrine glands |
| C76–C80 | Malignant neoplasms of indeterminate site, secondary site, and unknown site |
| C81–C96 | Malignant neoplasms of lymphoid, hematopoietic, and related tissues with a known or presumed primary |

| C97 | Primary multisite malignant neoplasms |
| D00–D09 | Intraepithelial neoplasm |
| D10–D36 | Benign new organism |
| D37–D48 | Neoplasms of unknown or unidentified nature |

A data set (number of items × number of time steps) was created for each patient using the following items: laboratory tests, gender, height and weight, and vital signs. Pre-processing of each item followed the method in Table 4. Data were ordered by date at the time of recording and treated as time-series data. The 4668 patients whose deaths were confirmed from the electronic medical record data were considered as dead patients, and the 8027 patients whose deaths were not confirmed were considered as surviving patients. Mortality information was not included in the input of the deep state-space model because it was used to interpret the results. For missing data, we used masked data indicating the presence or absence of missing data to account for missing data in the deep state-space model. The missing rate for the entire data set was 59.23%. The distribution of the number of time steps for each record per patient is shown in Figure 2 of expended data 3, with the number of patients on the vertical axis and the number of time steps on the horizontal axis. The highest number of time steps for the bottom 90% of patients in terms of time step count was 238 steps, and for patients with a higher number of time steps, the most recent record with 238 steps was used.

Table 4: Preprocessing methods

|  | Data to be use | Preprocessing methods | Number of items |
|---|---|---|---|
| Height and weight | Use daily mode | Normalization: MinMax method<br>Missing value interpolation: Zero-order spline method | 2 |
| Gender | Male: 1, Female: 0 | Enter the same gender for all time steps | 1 |
| Vitals | Use daily mode of body temperature, pulse, max. and min. blood pressure | Normalization: MinMax method<br>Missing value interpolation: Zero-order spline method | 4 |
| Laboratory tests | Use 50 items with the most abnormal values, except for correlation > 0.7<br>Abnormal high value: 1,<br>Abnormal low value: 0,<br>Normal value: 0.5 | Normalization: MinMax method<br>Missing value interpolation: Zero-order spline method | 50 |

**Deep state-space model**

In this study, latent state estimation was performed using the deep state-space model, a deep learning type state-space model, which can estimate the latent state of a time series by inputting the observed quantities of the time series, as shown in Figure 1b The layers and hyperparameters of the deep state-space model are described in expended data 4 of the Supplement.

In the deep state-space model, a neural network is used to represent a probabilistic state-space model, and the parameters of the three neural networks are learned from the observed data through variational inference. The state-space model here considers the observation $x_t$ from the state $z_t$ at each time t ($t = 1,2,...,T$) and represents the transformation from state to observation as $p_\theta(x_t|z_t)$ and the state transition as $p_\theta(z_t|z_{t-1})$. These probability distributions are represented using normal distributions parameterized by time-independent neural networks. These neural networks are trained by variational inference from unsupervised observed data only.

Variational inference introduces a new neural network-represented variational distribution $q_\phi$ for learning and learns the parameters of all neural networks by the gradient method by maximizing the lower bound of $\log p_\theta(x)$, which is the log-likelihood that is peripheral for the states derived as follows

$$\log p_\theta(x) \geq \sum_{t=1}^{T} E_{q_\phi(z_t|x)}[\log p_\theta(x_t|z_t)] - KL(q_\phi(z_1|x) \mid p_0(z_1))$$
$$- \sum_{t=2}^{T} E_{q_\phi(z_{t-1}|x)}[KL(q_\phi(z_t|z_{t-1},x) \mid p_\theta(z_t|z_{t-1}))] \quad (1)$$

where KL represents the KL divergence, defined by $KL(q(z) \mid p(z)) = \int_Z q(z) \log \frac{q(z)}{p(z)} dz$. The prior distribution $p_0(z_1)$ of the initial condition is assumed to be normally distributed with mean 0 variance 1. The first term represents the term related to the reproduction of observations, the second term represents the regularization term due to the prior of the initial distribution, and the third term represents the term related to the consistency of state transitions. Once these neural networks can be trained, the states can be estimated using $q_\phi(z_t|z_{t-1},x)$, which approximates the posterior distribution.

**Visualization of latent states**

Since the latent states are represented in multiple dimensions, we visualized them in two dimensions using UMAP, a dimensionality reduction technique, to facilitate their interpretation. the parameter selection for UMAP was performed as follows. First, the obtained latent states were down sampled to 1/10 and trials were made using the parameters n_neighbors= {15, 30, 50, 100} and

n_components=2. n_neighbors is a parameter that determines the width of the establishment distribution used in the establishment calculations for high-dimensional spaces. Lower values result in a microstructure, while larger values reflect a macrostructure in the dimensionality reduction results. Also, n_components is a parameter that represents the number of dimensions after dimensionality reduction. The default parameter setting of n_neighbors=15 was selected because the visualization results did not vary significantly by parameter. Since the latent states were obtained in time series, a transition diagram was obtained in two dimensions.

For this interpretation, a clustering method, the k-means method, was used to stratify patient status from latent status before dimensionality reduction. For the number of clusters, we selected clusters=3, which has a high silhouette score, a clustering evaluation index, and low variability in the total number of samples. In addition, we used the flag of death obtained from the electronic medical record data to interpret the latent state of patients immediately before their death as particularly critical. Therefore, we analyzed the clusters in which the last state in the time series of dead patients was concentrated as critical, and the clusters in which the last state in the time series of surviving patients was concentrated as stable.

**Latent state transition**

To investigate state transitions over time for patients, we considered transitions between clusters as Markov chains and calculated their transition probabilities for all patients', dead patients, and surviving patients' groups, respectively. Specifically, we counted the number of transitions between clusters in each group and compiled the frequencies into a table.

**Extract signals during state transitions**

For each cluster obtained by clustering latent states, we investigated the characteristic test items and extracted the test items that are important during latent state transitions. In each cluster, we investigated the laboratory test items in the input data that are considered abnormal values. Specifically, test values within the standard values used in the hospital were treated as normal values, those higher than the standard values as abnormal high values, and those lower than the standard values as abnormal low values. Laboratory items with many abnormal values are important factors during state transitions.

**Comparison Method**

PCA, VAE, and the linear state-space model were used as comparative methods for latent state estimation using the deep state-space model. PCA is a simple linear dimensionality reduction method that does not take time series into account, while VAE is a nonlinear dimensionality reduction method that does not take time series into account. VAE consists of two neural networks, an encoder and a decoder, with three all-connected layers for the encoder and two all-connected layers for the decoder.

As hyperparameters, we tried values of epoch=50 and learning rate= {0.001, 0.005, 0.01} and selected the parameter learning rate=0.01 that qualitatively captured the most mortality and survival states. The linear state-space model is a linear dimensionality reduction method that takes time series into account, specifically, it is a linearization of the neural networks that make up the deep state-space model. The 8-dimensional latent state obtained by applying the dimension reduction methods under the same conditions as the latent state estimation by the deep state-space model was visualized in two dimensions by UMAP. We compared whether the latent state obtained by each dimensionality reduction method adequately captures the patient's state, and whether it captures the patient's state transitions over time.

**Visualization and interpretation of the latent state of each injected cancer patient**

For eight drugs (nivolumab, trastuzumab, cisplatin, bicalutamide, imatinib, osimertinib, afatinib, and erlotinib), we visualized only the latent states of cancer patients treated with each anticancer drug and investigated the latent states, state transitions, and signals during state transitions that are characteristic of each drug.

## Conclusion

we developed a deep state-space analysis framework for estimating and visualizing latent states from EHR using the deep state-space model. We applied this framework to cancer patients treated with anticancer drugs and conducted a comprehensive analysis of the time-series changes leading up to death. The interpretation revealed three latent states and their relationship over time. The results identified potential important risk factors in cancer patients treated with anticancer drugs. Furthermore, the latent states obtained here were used for visualization and interpretation for cancer patients treated with specific anticancer drugs to obtain test items that are characteristic of cancer patients treated with specific anticancer drugs and are important during state transitions.

This framework is expected to lead to early treatment changes, prognostic decisions, and optimal treatment along the course of disease in long-term treatment strategies for cancer and other chronic diseases by understanding the latent state of disease progression from EHR. In addition, applying this framework to various high dimensional and complex time-series real-world data other than EHR can lead to the development of new time-series deep learning.

## References

[1] Rajkomar, Alvin, et al. "Scalable and accurate deep learning with electronic health records." *NPJ digital medicine* 1.1 (2018): 18.

[2] Shickel, Benjamin, et al. "Deep EHR: a survey of recent advances in deep learning techniques for electronic health record (EHR) analysis." *IEEE journal of biomedical and health informatics* 22.5 (2017): 1589-1604.

[3] Xiao, Cao, Edward Choi, and Jimeng Sun. "Opportunities and challenges in developing deep learning models using electronic health records data: a systematic review." *Journal of the American Medical Informatics Association* 25.10 (2018): 1419-1428.

[4] Rongali, Subendhu, et al. "Learning latent space representations to predict patient outcomes: Model development and validation." *Journal of Medical Internet Research* 22.3 (2020): e16374.

[5] Chushig-Muzo, David, et al. "Learning and visualizing chronic latent representations using electronic health records." *BioData Mining* 15.1 (2022): 1-27.

[6] Krishnan, Rahul G., Uri Shalit, and David Sontag. "Deep kalman filters." *arXiv preprint arXiv:1511.05121* (2015).

[7] McInnes, Leland, John Healy, and James Melville. "Umap: Uniform manifold approximation and projection for dimension reduction." arXiv preprint arXiv:1802.03426 (2018).

[8] Kao, Steven CH, et al. "High blood neutrophil-to-lymphocyte ratio is an indicator of poor prognosis in malignant mesothelioma patients undergoing systemic therapy." *Clinical cancer research* 16.23 (2010): 5805-5813.

[9] Ying, Hou-Qun, et al. "The prognostic value of preoperative NLR, d-NLR, PLR and LMR for predicting clinical outcome in surgical colorectal cancer patients." *Medical oncology* 31 (2014): 1-8.

[10] Liu, Zuqiang, et al. "Prognostic value of the CRP/Alb ratio, a novel inflammation-based score in pancreatic cancer." *Annals of surgical oncology* 24 (2017): 561-568.

[11] Li, Fan, et al. "Association between lactate dehydrogenase levels and oncologic outcomes in metastatic prostate cancer: A meta-analysis." *Cancer Medicine* 9.19 (2020): 7341-7351.

[12] Kumamoto, Y et al. "Clinical studies on endocrine therapy of prostatic carcinoma (1): Multivariate analyses of prognostic factors in patients with prostatic carcinoma given endocrine therapy". *Hinyokika kiyo. Acta urologica Japonica* vol. 36,3 (1990): 275-84.

## Supplements

**Expended data 1**

Table 1.1: Number of transitions between clusters for all patients (%)

|  |  | After transition |
|---|---|---|

|  |  | (1) | (2) | (3) |
|---|---|---|---|---|
| Before transition | (1) | 26.27 | 1.31 | 0.03 |
|  | (2) | 1.54 | 41.00 | 3.17 |
|  | (3) | 0.00 | 3.37 | 23.25 |

Table 1.2: Number of transitions between clusters of dead patients (%)

|  |  | After transition | | |
|---|---|---|---|---|
|  |  | (1) | (2) | (3) |
| Before transition | (1) | 36.45 | 1.43 | 0.02 |
|  | (2) | 1.85 | 38.21 | 2.49 |
|  | (3) | 0.00 | 2.80 | 16.71 |

Table 1.3: Number of transitions between clusters of surviving patients (%)

|  |  | After transition | | |
|---|---|---|---|---|
|  |  | (1) | (2) | (3) |
| Before transition | (1) | 19.22 | 1.24 | 0.03 |
|  | (2) | 1.33 | 42.93 | 3.65 |
|  | (3) | 0.00 | 3.77 | 27.78 |

**Expended data 2**

Table 2: Number of plots of final status of dead and surviving patients in each cluster for each cancer patient treated with anticancer drugs

Statistics displayed: Number of patients [persons]

|  |  | Cluster (1) | Cluster (2) | Cluster (3) | Sum |
|---|---|---|---|---|---|
| Nivolumab (404 persons) | Dead patients | 160 | 40 | 5 | 205 |
|  | Survive patients | 50 | 83 | 66 | 199 |
| Trastuzumab (267 persons) | Dead patients | 47 | 22 | 4 | 73 |
|  | Survive patients | 7 | 72 | 115 | 194 |
| Cisplatin (3799 persons) | Dead patients | 1275 | 426 | 74 | 1775 |
|  | Survive patients | 317 | 889 | 818 | 2024 |
| Bicalutamide (1064 persons) | Dead patients | 110 | 34 | 4 | 148 |
|  | Survive patients | 133 | 318 | 465 | 916 |
| Imatinib (125 persons) | Dead patients | 7 | 2 | 0 | 9 |
|  | Survive patients | 28 | 57 | 31 | 116 |
| Osimertinib | Dead patients | 13 | 7 | 1 | 21 |

| (103 persons) | Survive patients | 12 | 36 | 34 | 82 |
| Afatinib | Dead patients | 33 | 7 | 0 | 40 |
| (99 persons) | Survive patients | 11 | 22 | 26 | 59 |
| Erlotinib | Dead patients | 148 | 96 | 20 | 264 |
| (385 persons) | Survive patients | 21 | 64 | 36 | 121 |

The mean values of each test item in clusters (1) through (3) were investigated, and the 20 test items with the largest differences between the mean and normal values are shown in Figure 1.

Laboratory parameters specific to nivolumab-treated cancer patients were observed in cluster (1): high Segmented and low lymphocyte counts. Elevated levels of the tumor marker CEA were observed as a characteristic signal of transition to a dangerous state in trastuzumab-treated cancer patients. No characteristic laboratory tests were found in cisplatin-treated cancer patients compared to patients with cancer treated with all anticancer drugs. Characteristic signals of transition to dangerous states in cancer patients treated with bicalutamide were low levels of CK and high levels of LDH and BUN. Elevated CRE (serum creatinine) was found to be a characteristic signal of transition to a dangerous state in cancer patients treated with Imatinib. Characteristic signals of transition to danger status in cancer patients treated with osimertinib included high levels of Segmented (neutrophilic branch nuclei) and lymphocytes. In all clusters, high levels of D-dimer, tumor marker CYFRA and tumor marker CEA were also observed. Characteristic signals of transition to dangerous states in cancer patients treated with afatinib were observed, including high levels of LDH and BS (plasma glucose) and low levels of Band (rod-shaped nuclear cells). In all clusters, high levels of D-dimer, tumor marker CYFRA and tumor marker CEA, and low levels of Ca (calcium) were also observed. Characteristic signals of transition to dangerous states, such as high levels of LDH (lactate dehydrogenase), were observed in erlotinib-treated cancer patients. In all clusters, high levels of D-dimer, tumor marker CYFRA, and tumor marker CEA were also observed.

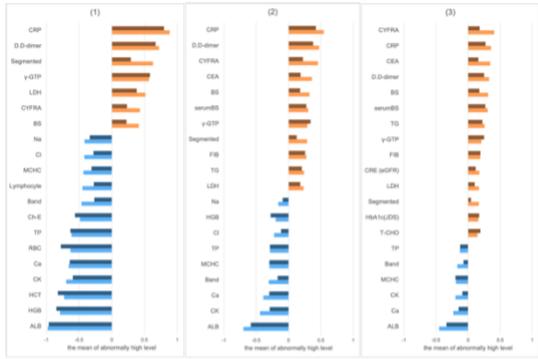
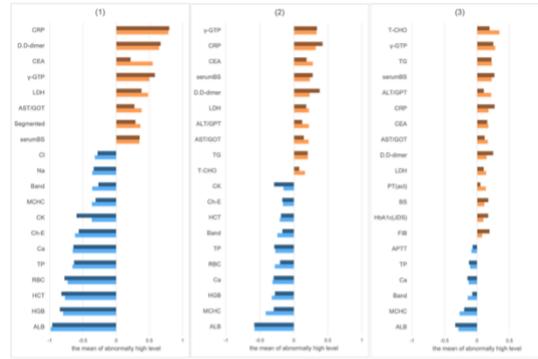
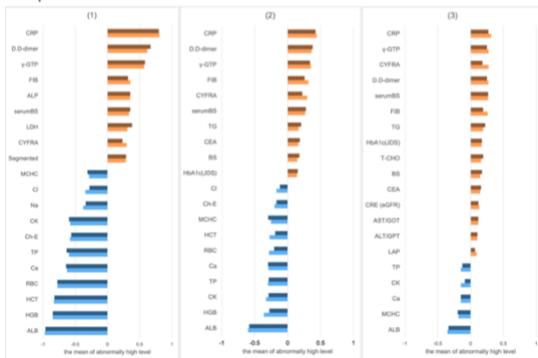
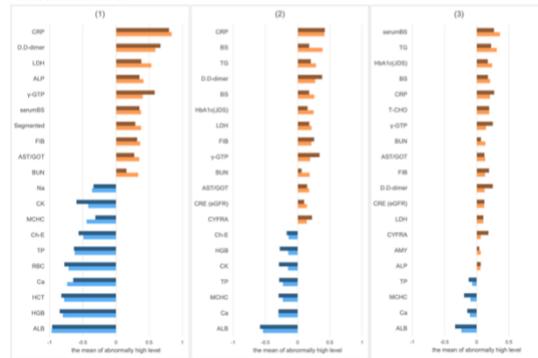
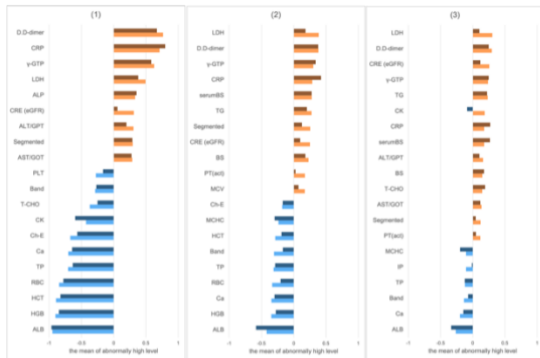
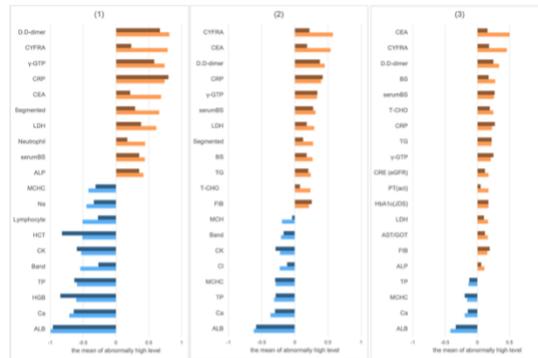
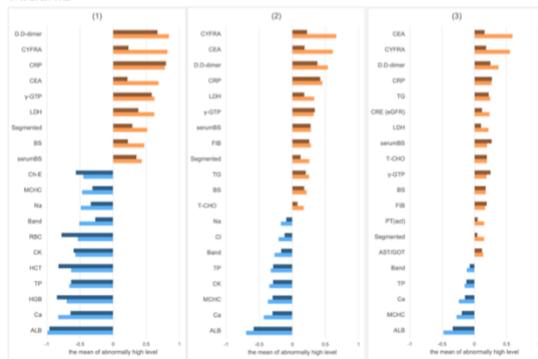
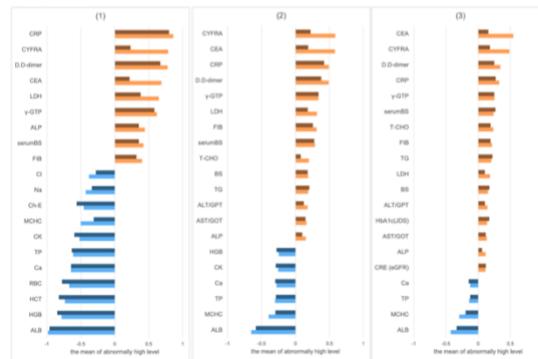

Figure 1: test items with large differences between the mean and normal values for each cluster in each administered cancer patient.

**Expended data 3**

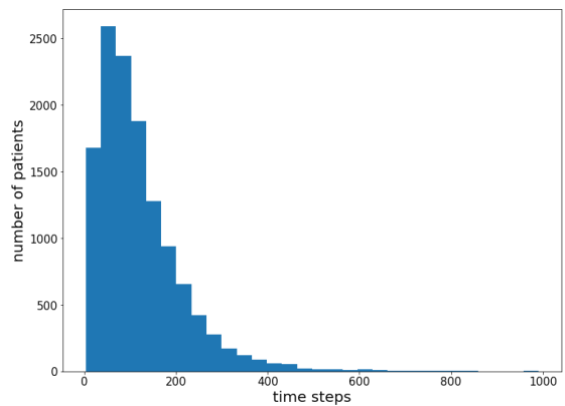

Figure 2: Number of time steps in records per patient

**Expended data 4**

Since the estimation results vary depending on the hyperparameters, we tried the parameters in Table 3 below, qualitatively capturing the mortality and survival states, and analyzed the results obtained with lr=0.005 and dim=8, which are the easiest to interpret. The neural network from state-space to observation consisted of two all-coupled layers, the neural network from state-space to potential used a dropout layer and an all-coupled layer, and the neural network from state-space (time t) to state-space (time t+1) used two all-coupled layers. In addition, the neural network from observation to state-space used two layers: an all-coupled layer and a long short-term memory (LSTM) layer.

Table 3: Parameter settings for deep state-space model

| Parameter name | Meaning | Parameter tried |
| --- | --- | --- |
| dim | Number of dimensions of latent state | 2, 4, 8, 16 |
| lr | Learning rate | 0.005, 0.01 |